\documentclass[color]{pbmlarxiv} 

\usepackage{xcolor}
\usepackage{amsmath}
\usepackage{cleveref}
\usepackage{booktabs}
\newcommand{\footnotehref}[2]{\footnote{\href{#1}{#2}}}

\newcommand{\fastalign}{fast\_align}
\begin{document}


\title{Leveraging Neural Machine Translation for Word Alignment}


\author{
  firstname=Vilém,
  surname=Zouhar,
  institute={uds},
  corresponding=yes,
  address={Institute of Formal and Applied Linguistics\\
    Faculty of Mathematics and Physics,\\
    Charles University\\
    Malostranské náměstí 25\\
    118 00 Praha 1, Czech Republic
  },
  email={vzouhar@lsv.uni-saarland.de},
}

\author{
  firstname=Daria,
  surname=Pylypenko,
  institute={uds},
  email={daria.pylypenko@uni-saarland.de},
}

\institute{uds}{Saarland University, Department of Language Science and Technology}

\shorttitle{Leveraging Neural MT for Word Alignment}
\shortauthor{V. Zouhar, D. Pylypenko}

\PBMLmaketitle

\begin{abstract}
The most common tools for word-alignment rely on a large amount of parallel sentences, which are then usually processed according to one of the IBM model algorithms. The training data is, however, the same as for machine translation (MT) systems, especially for neural MT (NMT), which itself is able to produce word-alignments using the trained attention heads. This is convenient because word-alignment is theoretically a viable byproduct of any attention-based NMT, which is also able to provide decoder scores for a translated sentence pair.

We summarize different approaches on how word-alignment can be extracted from alignment scores and then explore ways in which scores can be extracted from NMT, focusing on inferring the word-alignment scores based on output sentence and token probabilities. We compare this to the extraction of alignment scores from attention.
We conclude with aggregating all of the sources of alignment scores into a simple feed-forward network which achieves the best results when combined alignment extractors are used.
\end{abstract}

\section{Introduction}

Although word alignment found its use mainly in phrase-based machine translation (for generating phrase tables), it is still useful for many other tasks and applications: boosting neural MT performance \citep{alkhouli2016alignment}, exploring cross-linguistic phenomena \citep{schrader2006does}, computing quality estimation \citep{specia2013quest}, presenting quality estimation \citep{zouhar2020extending} or simply highlighting matching words and phrases in interactive MT (publicly available MT services).

The aim of this paper is to improve the word alignment quality and demonstrate the capabilities of alignment based on NMT confidence. Closely related to this is the section devoted to aggregating multiple NMT-based alignment models together, which outperforms the individual models.
This is of practical use (better alignment) as well as of theoretical interest (word alignment information encoded in NMT scores). 

We first briefly present the task of word alignment, the metric and the used tools and datasets.
In \Cref{sec:individual} we introduce the soft word alignment models based on MT scores and also several hard word alignment methods (extractors).
The models are evaluated together with other solutions (\fastalign{} and Attention) in \Cref{sec:evaluation}.
We then evaluate the models enhanced with new features and combined together using a simple feed-forward neural network (\Cref{sec:aggregated}).
In both cases, we explore the models' behaviour on Czech-English and German-English datasets.

All of the code is available open-source.\footnotehref{https://github.com/zouharvi/LeverageAlign}{github.com/zouharvi/LeverageAlign}

\subsection{Word Alignment} \label{subsec:word_alignment}

\begin{figure*}[ht!]
    \vspace{0.01cm}
    \center
    \includegraphics[width=0.87\textwidth]{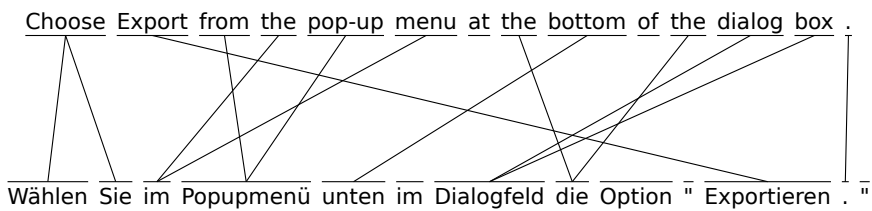}
    \caption{
        Illustration of English (top) to German (bottom) word alignment. The token >>choose<< is aligned to two tokens >>wählen<< and >>Sie<< while the token >>Option<< is left unaligned. The article >>die<< is mistakenly aligned to two unrelated articles >>the<<.
        \label{fig:alignment_example}
    }
    \vspace{0.2cm}
\end{figure*}

Word alignment (also bitext alignment) is a task of matching two groups of words together that are each other's semantic translation. This is problematic for non-content words which are specific for the given language but generally one is able to construct a mapping as in the example in \Cref{fig:alignment_example}. Word alignment usually follows after sentence alignment.
Even though it is called word alignment, it usually operates on all tokens, including punctuation marks.

Word alignment output can be formalized as a set containing tuples of source-target words. For an aligner output $A$, a sure alignment $S$ and a possible alignment $P$ ($S\subseteq P$),\footnote{Sure alignments can be treated as gold alignments with very high confidence, while pairs marked with possible alignments are still sensible to connect, but with the decision being much less clear. The AER is designed not to penalize models by including more possible alignments in the gold annotations.} precision can be computed as $ \frac{|A \cap P|}{|A|}$ and recall as $\frac{|A \cap S|}{|S|}$. The most common metric, Alignment Error Rate (AER), is defined as $1-\frac{|A\cap S| + |A \cap P|}{|S|+|A|}$ (lower is better). Even though the test set is annotated with two types of alignments, the aligner is expected to produce only one type. These evaluation measures are described by \citet{mihalcea2003evaluation} and \citet{och2003systematic}.

Traditionally word alignment models can be split into soft and hard alignment parts. In soft alignment, the model produces a score for every source-target pair. When producing hard alignment (extractors), the model makes decisions as to which alignments to include in the output. For source sentence $S$ and target sentence $T$, the output of soft alignment is a $\mathbb{R}^{|S|\times|T|}$ matrix while hard alignment is a set $A \subseteq S\times T$.

\paragraph{Symmetrization.}

Assuming that we have access to bi-directional word alignment (in the context of this paper to two MT systems of the opposite directions) we can compute both the alignment from source to target ($X$) and target to source ($X'$).
Having access to both $X$ and $X'$ makes it possible to create a new alignment $Y$ with either higher precision through intersection or higher recall through union \citep{koehn2009statistical}.
\begin{gather*}
    X^T := \{(b, a): (a, b) \in X\} \\
    Y_{prec} = X \cap X'^T \qquad Y_{rec} = X \cup X'^T
\end{gather*}

We can make use of the fact that the models output soft alignment scores and create new alignment scores in the following way using a simple linear regression model. This allows us to fine-tune the relevance of each of the directions as well as their interaction. However, it does not have the same effects as the union or the intersection because it affects the soft alignment and not hard alignment in contrast to the previous case.
\begin{gather*}
    p^{sym}(s,t) = \beta_0 \cdot p(s, t) + \beta_1 \cdot p^r(t, s) + \beta_2 \cdot p(s, t) \cdot p^r(t, s)
\end{gather*}

More complex symmetrization techniques have been proposed and implemented by \citet{och2000improved, junczys2011symgiza++}.

\subsection{Relevant Work}

\citet{och2003systematic} introduce the word alignment task and systematically compare the IBM word alignment models.
The work of \citet{li2019word} is closely related to this article as it examines the issue of word alignment from NMT and proposes two ways of extracting it: prediction difference and explicit model. They also show that without guided alignment in training, NMT systems perform worse than \fastalign{} baseline.
Using attention for word alignment is thoroughly discussed by \citet{bahdanau2014neural} and \citet{zenkel2019adding}.
Word alignment based on static and contextualized word embeddings is explored by the recent work of \citet{sabet2020simalign}.
Word alignment based on cross-lingual (more than 2 languages) methods is presented by \citet{wu2021slua}.
The work of \citet{chen2020accurate} focuses on inducing word alignments from glass-box NMT as a replacement for using Transformer attention layers directly.
\citet{chen2020mask} document Mask Align, an unsupervised neural word aligner based on a single masked token prediction.

\citet{chen2016guided} propose guided attention, a mechanism that uses word alignment to bias the attention during training. This improves the MT performance on especially rare and unknown tokens. The usage of word alignment in this work is, however, opposite to the goals of this paper. While for \citet{chen2016guided} the word alignment improved their MT system, here the MT system improves the word alignment.

\subsection{Tools}

The experiments in this paper require an MT system capable of providing output probabilities (decoder scores) and optionally also attention-based word alignment. For comparison, we also use an IBM-model-based word aligner. This tool is also used as an additional feature to the final aggregation model.

\paragraph{MarianNMT}\hspace{-0.25cm} \citep{junczys2018marian, junczys2018marian2} is a popular (both in academia and in deployment scenarios), actively developed and maintained system for fast machine translation. It already contains options for producing word alignment, output probabilities for words and sentences and also attention scores.

\paragraph{\fastalign}\hspace{-0.25cm} \citep{dyer2013simple} is an unsupervised word aligner based on IBM Alignment Model 2. It does not provide state of the art pre-neural performance but is easy to build with modern toolchains and has low resource requirements (both memory- and computational-wise).

\subsection{Data}

For training purposes, we make use mostly of the parallel corpora of Czech--English word alignments by \citet{marecek_csen_algn_corpus}, based on manually annotated data. We also include a large Czech-English corpus by \citet{czeng2} and a large German--English corpus by \citet{rozis_tilde}, which are not word aligned. From this corpus, 1M sentences were sampled randomly. A small manually aligned German--English corpus by \citet{bicini_ende_algn_corpus} is included for testing. An overview of the corpus sizes is displayed in \Cref{tab:corpus_used}.

\begin{table*}[h!]
    \center
    \begin{tabular}{lccrrr}
        \toprule
        CS/DE-English & Type &\hspace{-0.1cm}Domain &\hspace{-0.1cm}CS/DE Tok. & EN Tok. & Sent. \\
        \cmidrule{1-6}
        Czech Small & aligned & news, legal & 53k & 60k & 2.5k \\
        Czech Big & unaligned & multi & 2618M & 3013M & 188M \\
    German Small\hspace{-0.2cm} & aligned & legal & 1k & 1k & 0.1k \\
        German Big & unaligned &tech, news, legal & 23M & 25M & 1M \\
        \bottomrule
    \end{tabular}
    \caption{Used word aligned corpora with their sizes, domains and origin. \label{tab:corpus_used}}
\end{table*}

\subsection{MT Models}

We make use of the MT models made available\footnotehref{https://github.com/browsermt/students}{github.com/browsermt/students} by \citet{model_csen} and \citet{model_deen}. For both Czech-English and English-German, CPU-optimized student models are used. They are transformer-based \citep{vaswani2017transformer} and were created by using knowledge distillation. With WMT19 and WMT20 SacreBLEU \citep{sacrebleu}, the models achieve the following BLEU scores: Czech-English (27.7), English-Czech (36.3) and English--German (42.7).\footnote{%
BLEU+case.mixed+lang.cs-en+numrefs.1 +smooth.exp+test.wmt20+tok.13a+version.1.4.13\\
\hspace*{0.49cm}BLEU+case.mixed+lang.en-cs+numrefs.1 +smooth.exp+test.wmt20+tok.13a+version.1.4.13\\
\hspace*{0.49cm}BLEU+case.mixed+lang.en-de+numrefs.1+smooth.exp+test.wmt19+tok.13a+version.1.4.8} Since the English--German MT is only available in one direction, word alignment is reported in this direction as well. Exceptions, such as word alignment using an MT for the opposite direction, are explicitly mentioned.
\section{Individual Models} \label{sec:individual}

In this section, we describe and evaluate the individual word alignment models. All of the newly introduced models make use of the fact that NMT systems can be viewed as language models and can produce translation probabilities.

\subsection{Baseline Models} \label{subsec:baseline_models}

The first model is \fastalign{}. The second is attention-based soft word alignment extracted from MarianNMT (Attention), which was trained with guided alignment during the distillation. For the rest of this subsection, we will focus on models generating soft alignment scores (an unbounded real number corresponding to the quality of a possible alignment between two tokens) and not the alignments themselves.

\paragraph{One Token Translation ($M_1$).}

The simplest approach to get alignment scores is to compute decoder translation probability using the MT (function $m$) between every source and target token $s_i$ and $t_j$ of the source and target sentences $S$ and $T$. Single tokens are passed to the models as if they were a sentence pair. The scores are not normalized which is not an issue in this case, since the models working with these alignment scores (in \Cref{subsec:extractors}) compare output from sequences of the same length.
\begin{gather*}
\forall s_i\in S, t_j \in T: p(s_i, t_j) = m(\{s_i\}, \{t_j\})
\end{gather*}

The produced values are in a log space $(-\inf, 0]$. This approach requires $|S|\cdot |T|$ of one-token translation scorings (decoder probability of the target reference) for producing word alignments of a single sentence pair. On a CPU,\footnote{8 threads 2.3GHz Ryzen 7 3700u, no RAM to disk swapping} the models average to $2.7$s per one sentence pair alignment.

\paragraph{Source Token Dropout ($M_2$).}

A more refined approach was chosen by \citet{zintgraf2017visualizing} in which the alignment score is computed as the difference in target token probability when the source token is unknown. The exact approach is too computationally demanding (requires translation scorings with large amounts of replacement words), and therefore we use a much simpler, yet conceptually similar method by either omitting the token or replacing it with \texttt{<unk>}.\footnote{Even though subword-based MT models do not need \texttt{<unk>}, SentencePiece reserves the token \texttt{<unk>} for an unknown symbol.}
Assume $m_j(S, T)$ produces the log probability of the $j$-th target token. The sentence $S^{a/b}_i$ with an obscured token $s_i$ can be defined in two ways which leads to two versions of this model: $M_2^a$ and $M_2^b$. Output is then possibly unbounded $(-\inf,\inf)$.
\begin{gather*}
    \forall s_i \in S, t_j \in T: p(s_i, t_j) = m_j(S, T) - m_j(S^{a/b}_i, T)
\end{gather*}

\vspace{-0.8cm}

\begin{align*}
    & \text{Word deletion}\, (M_2^a): 
    & S^{a}_i = s_0, s_1, \ldots, s_{i-1}, s_{i+1}, \ldots, s_{|S|} \\
    & \text{Word substitution}\, (M_2^b):
    & S^{b}_i = s_0, s_1, \ldots, s_{i-1}, \texttt{<unk>}, s_{i+1}, \ldots, s_{|S|}
\end{align*}

This requires $|S|$ translation scorings of source and target lengths $|S|$ and $|T|$, which is comparable to $M_1$. The models average to $1.5$s per one sentence pair alignment.\footnote{ The running time is lower because in this case it is $|S|$ scorings of length $|T|$, while in $M_1$ it is $|S|\times|T|$ scorings of length $1$.}

\paragraph{Source and Target Dropout ($M_3$).}

A very similar method would be to also dropout the target token and examine how the sentence probability changes. Applying the two different ways of dropout leads to four versions of this approach. Note that in this case we compute the sentence probability (because the target word is hidden) and also do not subtract from the base sentence probability, but rather use the new sentence probability as it is. This probability should be highest if the corresponding tokens are both obscured. The probability is in log space $(-\inf, 0]$.
\vspace{0.2cm}
\begin{gather*}
    \forall s_i \in S, t_j \in T: p(s_i, t_j) = m(S^{a/b}_i , T^{a/b}_j)
\end{gather*}

\vspace{-0.8cm}

\begin{align*}
    T^a_j &= t_0, t_1, \ldots, t_{j-1}, t_{j+1}, \ldots, t_{|T|} \\
    T^b_j &= t_0, t_1, \ldots, t_{j-1}, \texttt{<unk>}, t_{j+1}, \ldots, t_{|T|}
\end{align*}

\vspace{-0.8cm}

\begin{align*}
    & \text{Word deletion, deletion ($M_3^{aa}$)} & S^a_i, T^a_j \\
    & \text{Word deletion, substitution ($M_3^{ab}$)} & S^a_i, T^b_j \\
    & \text{Word substitution, deletion ($M_3^{ba}$)} & S^b_i, T^a_j \\  
    & \text{Word substitution, substitution ($M_3^{bb}$)} & S^b_i, T^b_j
\end{align*}

This approach requires $|S|\cdot|T|$ translation scorings of source and target lengths of $|S^{a/b}|$ and $|T^{a/b}|$ for sentence $S$ translated to $T$ which is roughly $|T|$ times more than in $M_1$ and $M_2$. This makes it it the most computationally demanding approach. On average it takes $46.1$s to produce one sentence pair alignment on a CPU.

\subsection{Direct Alignment from Baseline Models} \label{subsec:extractors}

All of the models (except for \fastalign{}) are not producing the alignments themselves, but soft alignment scores $p$ for each pair of tokens $(s, t)$ in source $S$ $\times$ target $T$ sentence. The hard alignment itself can then, for example, be computed in the following ways. The parameter $\alpha$ can be estimated from the development set. The function $p$ is in general any soft alignment function (e.g. attention scores or the alignment scores from IBM model 1 EM algorithm).

\begin{enumerate}
    \item For every source token $s$ take the target tokens $t$ with the maximum score.
    \begin{gather*}
        A_1 = \bigcup_{s \in S} \{ (s, t): p(s,t) = max_r \{p(s,r) \} \}
    \end{gather*}
    
    \item For every source token $s$ take all target tokens $t$ with a high enough score (above threshold). This method is used to control the density of alignments in the work of \citet{liang2006alignment} and provides a parameter to tradeoff precision and recall.
    \begin{gather*}
        A_2^\alpha = \bigcup_{s \in S} \{(s, t): p(s,t) \ge \alpha \}
    \end{gather*}
    
    \item For every source token $s$ take any target token which has a score of at least $\alpha$ times the score of the best candidate. Special handling for negative cases in the form of a division is required to make the formula work for the whole $\mathbb{R}$. The motivation for this is $M_2$, which provides possibly unbounded alignment scores. Assume $\alpha \in (0, 1]$.
    \begin{gather*}
        A_3^\alpha = \bigcup_{s \in S} \{ (s, t): p(s,t) \ge min
        \big[ \max_r\ p(s,r) \cdot \alpha, \max_r\ p(s,r)\, /\, \alpha \big] \}
    \end{gather*}
    $A_1$ can then be expressed as $A_3^1$. Lower $alpha$ values lead to lower precision and higher recall because the algorithm includes more, less probable, alignments. A variation on this approach would be to subtract $\alpha$ instead of multiplying it. The reason for choosing multiplication is that it dynamically adapts to a wider range of intervals and bounds the parameters between $0$ and $1$. This is not the case for substraction and because of this, it would be harder to choose the right parameter. 
    
    \smallskip
    \item Similar approach is for $A_3$, but with the focus on the target side. For every target token $t$ take any source token which has a score of at least $\alpha$ times the score of the best candidate.
    \begin{gather*}
        A_4^\alpha = \bigcup_{t \in T} \{ (s, t): p(s,t) \ge min
        \big[ \max_r\ p(r,t) \cdot \alpha, \max_r\ p(r,t) / \alpha \big] \}
    \end{gather*}
    Similar reversal for $A_2$ would not make sense, because it already takes all alignment above a threshold without any consideration for the direction.
    
    \item Similarly to $M_3$ and $M_4$ it is possible to create an extractor in which instead of having a single dropout on the target side, there are a multiple of them. This way, the score would not be between the source token and the target token, but between the source token and a subset of all target tokens. Formally, this would replace the (complete) weighted graph structure with a (complete) hypergraph. Instead of just having a weight for \textit{Choose--Wählen}, there would also be a weight for \textit{\{Choose\}--\{Wählen, Sie\}}, \textit{\{Choose\}-\{Wählen, im, Popupmenü\}} etc. 
    This would, however, lead to exponential complexity in terms of target sentence length. The number of words participating in an edge would then have to be limited to the number of alignments to a single token that we can empirically expect of the given language pair. \Cref{fig:alignment_example} suggests that for English-German this could be 3.
    Upon computing the scores for all the edges in this hypergraph, a follow-up task would be to find the maximum-weight matching.
\end{enumerate}

\vspace{-0.4cm}
\paragraph{Coverage.}

\begin{figure}[h!]
    \center
    \includegraphics[width=0.5\textwidth]{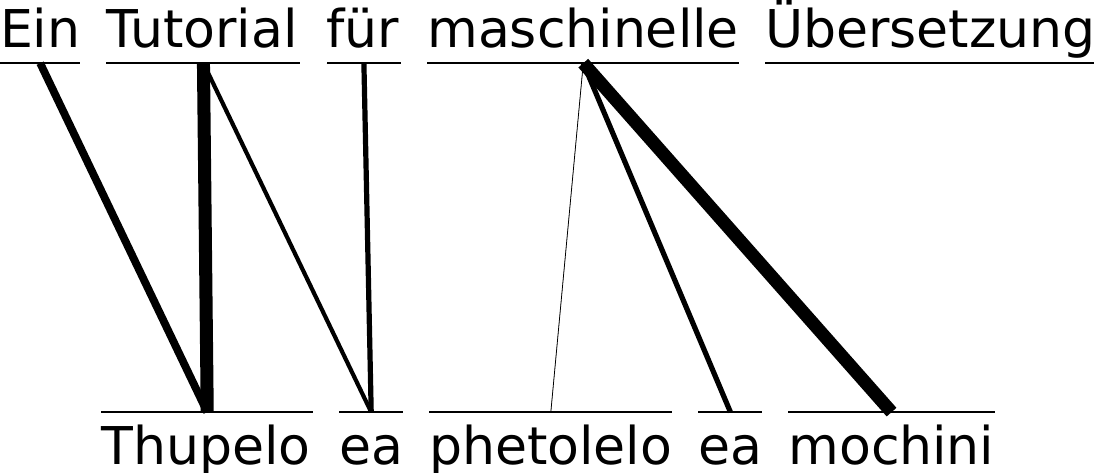}
    \caption{
        Partial alignment from German (top) to Sesotho (bottom). The model has no lexical knowledge about the alignment of >>Übersetzung<<, though >>phetolelo<< is a good candidate because no other word aligned to it. Line strength corresponds to the soft alignments produced by the model.
        \label{fig:alignment_coverage}
    }
\end{figure}

The suggested greedy way of computing alignments from alignment scores is far from perfect.
In the scenario depicted in \Cref{fig:alignment_coverage}, all but the last source token (German) have been aligned with the target, each with different alignment scores. Although the model may lack any lexical knowledge of the word \textit{Übersetzung}, it should consider the prior of a word being aligned to at least one target token.

In this specific case, $A_3^{0.9}$ would probably include all alignments to the word \textit{Übersetzung}, since there is no single strong candidate (assume that lines not visible depict soft alignments close to $0$). Similarly, $A_4^{0.9}$ would also include most alignments of the word \textit{Übersetzung}, including the word \textit{phetolelo}, since the alignment score with \textit{maschinelle} is weak and also close to $0$. Intersecting these two extractors $A_3^{0.9} \cap A_4^{0.9}$ would yield the correct alignment \textit{Übersetzung--phetolelo}. Other tokens would not be aligned to either of these two words because they have strong alignment scores with different tokens.

This prior may not always be desirable. For this, intersecting with $A_2^\alpha$ provides a limiting threshold. In an application where the target token is erroneous, this prevents the alignment model from aligning the two corresponding tokens.
Inducing alignment based on graph properties is examined by \citet{symmetric}, though without the presence of NMT.
\section{Evaluation of Individual Models} \label{sec:evaluation}

\begin{figure*}[h!]
    \center
    \includegraphics[width=0.92\textwidth]{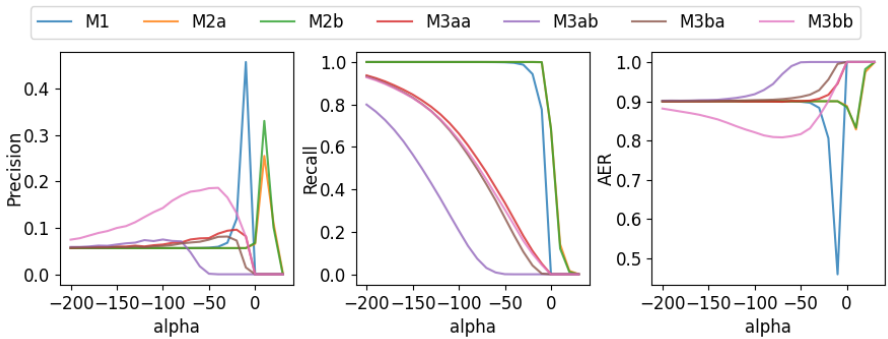}
    \vspace*{-0.3cm}
    \caption{Precision, Recall and AER of individual models on CS$\leftrightarrow$EN data extracted using $A_2$ (directions averaged) \label{fig:individual_encs_mix_a2}}
\end{figure*}

\paragraph{Baseline Models.} \Cref{fig:individual_encs_mix_a2} shows the results on Czech$\leftrightarrow$English data averaged from both directions. Different models have different spans of their scores, and therefore it is much harder to select the single best $\alpha$. The most basic model, $M_1$, achieves the best performance (AER = $0.46$). The figure serves as an illustration of the $A_2^\alpha$ landscape.

\begin{figure*}[h!]
    \center
    \includegraphics[width=0.93\textwidth]{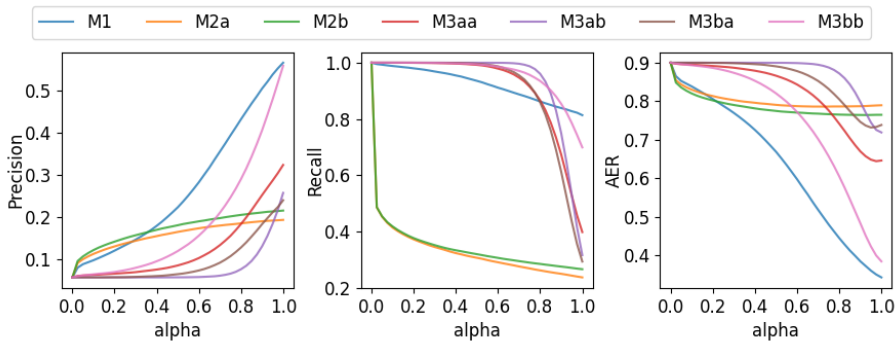}
    \vspace*{-0.4cm}
    \caption{Precision, Recall and AER of individual models on CS$\leftrightarrow$EN extracted using $A_3$ (directions averaged) \label{fig:individual_encs_mix_a3}}
    \vspace*{0.15cm}
\end{figure*}

The results on Czech$\leftrightarrow$English data averaged from both directions with $A_3$ can be seen in \Cref{fig:individual_encs_mix_a3}. The case of $\alpha = 0$ corresponds to aligning everything with everything, while $\alpha = 1$ means aligning only the token with the highest score to the single source one (i.e. $A_1$). The different model families behave similarly with respect to Precision, Recall and AER. $M_1$ achieves again the best result (AER = $0.34$), but with a smoother distinction between models.

\begin{figure*}[h!]
    \center
    \includegraphics[width=0.93\textwidth]{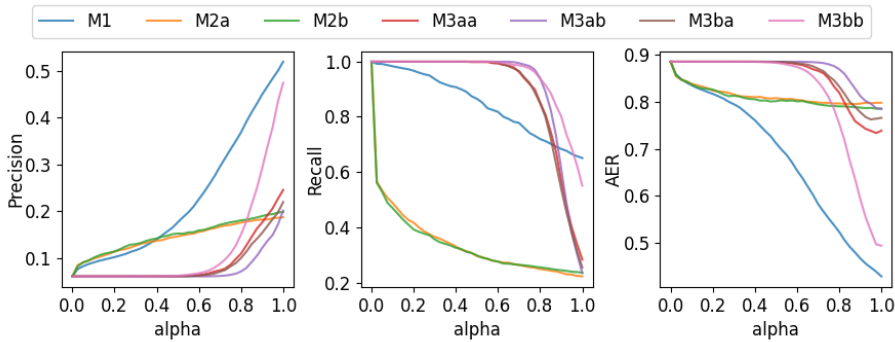}
    \vspace*{-0.4cm}
    \caption{Precision, Recall and AER of individual models on EN$\rightarrow$DE extracted using $A_3$ \label{fig:individual_ende_a3}}
    \vspace*{0.15cm}
\end{figure*}

Out of the model $M_3$ family, $M_3^{bb}$ outperformed the rest significantly. In $A_2$ (\Cref{fig:individual_encs_mix_a2}), the other models, $M_3^{aa}$, $M_3^{ab}$ and $M_3^{ba}$, perform worse than $M_2^a$ and $M_2^b$. This is reversed in case of using the $A_3$ extractor, as shown in \Cref{fig:individual_encs_mix_a3} and \Cref{fig:individual_ende_a3}. For the $M_3$ model family, models with mixed obscuring functions ($M_3^{ab}$ and $M_3^{ba}$) perform worse than with the same obscuring function on both the source and the target side ($M_3^{aa}$ and $M_3^{bb}$).

The English$\rightarrow$German dataset proved to be more difficult. The AER, that are shown in \Cref{fig:individual_ende_a3}, are higher than for Czech$\leftrightarrow$English. The model $M_1$ again achieves the best results with AER = $0.43$. The model ordering is preserved from \Cref{fig:individual_encs_mix_a3}.

\begin{figure*}[h!]
    \center
    \includegraphics[width=1.0\textwidth]{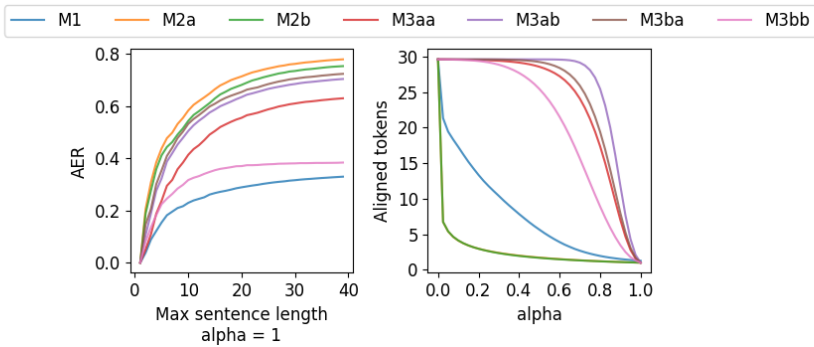}
    \vspace*{-0.4cm}
    \caption{AER for $\alpha=1$ (left) and average number of aligned tokens (right) of individual baseline models on CS$\leftrightarrow$EN extracted using $A_3$ (directions averaged) \label{fig:individual_ts_encs_mix_a3}}
    \vspace*{-0.2cm}
\end{figure*}

\Cref{fig:individual_ts_encs_mix_a3} documents that different model types produce different number of alignments per one token. It also shows that the performance rapidly decreases with sentence length. The high AER in \Cref{fig:individual_encs_mix_a3} can be explained by the dataset containing mostly longer sentences (21 tokens on average). The model $M_1$ is still better than $M_3^{bb}$ even on longer sentences despite the fact it does not model the context.

The best results were achieved with $A_4^1$ using $M_1$: AER = $0.30$ for German$\rightarrow$English and AER = $0.31$ for Czech$\leftrightarrow$English. The plots (not shown) are very similar to those of $A_3$. Hence $M_3^{bb}$ follows up with AER = $0.38$ and AER = $0.36$ for German and Czech respectively using $A_4^1$.

\begin{table*}[h!]
    \center
    \begin{tabular}{lccc}
        \toprule
        Data & Precision & Recall & AER \\
        \midrule
        Czech$\leftrightarrow$English Small & $0.54$ & $0.66$ & $0.41$ \\
        Czech$\leftrightarrow$English Big & $0.63$ & $0.64$ & $0.38$ \\
        German$\rightarrow$English Small & $0.49$ & $0.55$ & $0.48$\\
        German$\rightarrow$English Small+Big & $0.63$ & $0.72$ & $0.34$ \\
        \bottomrule
    \end{tabular}
    \caption{Precision, Recall and AER of \fastalign{}. Models were evaluated on the respective annotated datasets part. \label{tab:individual_fastalign}}
\end{table*}

\paragraph{\fastalign{}.} For comparison, the results of \fastalign{} can be seen in \Cref{tab:individual_fastalign}. For both language pairs, we use two models, trained on the Small and Big corpora. The motivation for the latter is that the performance of \fastalign{} on 5k sentence pairs is unfairly low in comparison to the other methods because the used MT system has had access to a much larger amount of data. This is shown by the performance difference between these two models.

\begin{table*}[h!]
    \vspace{0.5cm}
    \center
    \begin{tabular}{lcccc}
        \toprule
        Data & Subword Aggregation & Precision & Recall & AER \\
        \midrule
        Czech$\leftrightarrow$English Small & maximum & $0.64$ & $0.81$ & $0.29$ \\
        Czech$\leftrightarrow$English Small & average & $0.64$ & $0.81$ & $0.29$ \\
        German$\rightarrow$English Small\hspace*{-0.5cm}& maximum & $0.69$ & $0.81$ & $0.26$ \\
        German$\rightarrow$English Small\hspace*{-0.5cm}& average & $0.68$ & $0.80$ & $0.27$ \\
        \bottomrule
    \end{tabular}
    \caption{Precision, Recall and AER of attention-based word alignment extracted using $A_3^1$ \label{tab:individual_marian}}
\end{table*}

\paragraph{Attention Scores.} Extracting alignment from MT model attention using $A_3^1$ results in the highest performance (\Cref{tab:individual_marian}). Since the attention scores are between subword units from SentencePiece \citep{kudo2018sentencepiece}, we chose two methods of aggregation to a single score between two tokens (two lists of subwords): (1) taking the maximum probability between two subwords and (2) taking the average probability. They, however, produce almost identical results with respect to the word alignment quality. Scores are listed with $A_3^1$, but $A_2^{0.25}$ achieved very close results.

\begin{table}[h!]
    \vspace{0.5cm}
    \center
    \begin{tabular}{lcccc}
        \toprule
        Model & Method & Precision & Recall & AER \\
        \midrule
        $M_1$ & reverse & $0.56$ & $0.82$ & $0.35$ \\
        $M_1$ & add & $0.59$ & $0.86$ & $0.31$ \\
        $M_1$ & intersect & $0.73$ & $0.77$ & $0.26$ \\
        \midrule
        Attention (avg) & reverse & $0.64$ & $0.81$ & $0.29$ \\
        Attention (avg) & multiply & $0.66$ & $0.83$ & $0.28$ \\
        Attention (avg) & intersect & $0.77$ & $0.70$ & $0.28$ \\
        \bottomrule
    \end{tabular}
    \caption{Average Precision, Recall and AER on Czech$\leftrightarrow$English extracted using $A_4^1$ with symmetrization methods applied for $M_1$ and Attention (avg)\label{tab:individual_symmetry}}
\end{table}

\paragraph{Symmetrization.} Results of symmetrization methods (akin to those described in \Cref{subsec:word_alignment}) for $M_1$ and Attention scores (attention scores aggregated by averaging) are shown in \Cref{tab:individual_symmetry}. Each method is accompanied by an example formula; $p^x$ stands for either $M_1$ or Attention (avg) (in principle any function which produces soft alignments). Similarly, $A_4^1$ could be replaced by other extractors, even though this one worked the best. For \textit{reverse} and \textit{add}, $A_4^1$ is applied on the final result, but for simplicity left out of the formulas.

Method \textit{reverse} consists of using TGT$\rightarrow$SRC translation direction to get alignment scores but then transposing the soft alignment matrix so that the scores are SRC$\rightarrow$TGT.

\vspace*{-0.3cm}
\begin{gather*}
    p_{CS\rightarrow EN}^\text{reverse}(s, t) = p^x_{EN\rightarrow CS}(t, s)
\end{gather*}
\vspace*{0.0cm}

Method \textit{add} simply combines the original and reversed scores before alignments are extracted. The scores of $M_1$ are in log space; therefore, addition is used instead of multiplication. For attentions, multiplication is used, since they are bounded by $[0,1]$.

\vspace*{-0.3cm}
\begin{gather*}
    p_{CS\rightarrow EN}^\text{add}(s, t) = p^x_{CS\rightarrow EN}(s, t)+p^x_{EN\rightarrow CS}(t, s) \\
    p_{CS\rightarrow EN}^\text{mutliply}(s, t) = p^x_{CS\rightarrow EN}(s, t)\cdot p^x_{EN\rightarrow CS}(t, s)\hspace{0.17cm}
\end{gather*}
\vspace*{0.0cm}

Method \textit{intersect} first extracts the alignments for the two directions and then intersects the results (with one direction transposed). This method produces the best results overall (AER = $0.26$), also surpassing $M_1$'s forward direction and attention-based alignments.

\vspace*{-0.3cm}
\begin{gather*}
    A_4^1(p_{CS\rightarrow EN}^\text{intersect}(s, t)) = A_4^1(p^x_{CS\rightarrow EN}(s, t))\cap A_4^1(p^x_{EN\rightarrow CS}(t, s))
\end{gather*}
\vspace*{0.0cm}

In contrast to $M_1$, none of the other models, including attention-based, improved rapidly. This is partly explained by the fact that in other models, the precision-recall balance is shifted from recall to precision, while in $M_1$ it became more balanced after intersection. The reversal also allowed us to get significant results (AER = $0.27$) for the English$\rightarrow$German direction using Attention (avg), for which we did not have an MT system.

\subsection{Extractor Limitations}

Computing word alignments by taking the most probable target token ($A_3^1, A_4^1$) has theoretical limitations to the AER because it makes a faulty assumption that every token is aligned to at least one other token. The Czech$\rightarrow$English dataset has $12\%$ of unaligned tokens and an average of $1.16$ aligned target tokens per source tokens (excluding non-aligned tokens). 

Assuming access to a word alignment oracle ($0$ if not aligned, $1$ if aligned), in case the token is not aligned to any other, all of the scores are $0$. The extractor $A_3^1 = A_1$ will then take all tokens with values equal to the maximum, effectively aligning the in reality unaligned token to every possible one. This extractor is then bound to have maximum recall, but relatively poor accuracy.

The measured performance shows that the $A_2^{\alpha}$ is not the best extraction method. However, it is objectively not prone to this issue because it does not make any assumptions about the number of aligned tokens, and the minimum possible AER is 0 ($A_2^1$ with an oracle). In the next section, we will therefore make use of $A_2^{\alpha_1} \cap A_3^{\alpha_2} \cap A_4^{\alpha_3}$, which provides better performance than individual extractors.
\section{Ensembling of Individual Models} \label{sec:aggregated}

In the previous section, we saw that multiple methods with different properties achieved good results, but were sensitive to the method used to induce hard alignment. This section combines them together in a small feed-forward neural network, which can be trained on a small amount of data.

\subsection{Model}

The ensemble neural network itself is a regressor: $\mathbf{F} \rightarrow (0, 1)$, where $\mathbf{F}$ is the set of feature vectors for every pair of source and target tokens.\footnote{A completely different approach would be to simply use (pretrained) word embeddings as an input to the network. This is, however, not possible due to the low amount of gold alignment data.} By applying sigmoid to the output and establishing a threshold value for the positive class, the network would become a classifier. This behaviour can, however, be simulated using $A_2^\alpha$. We work with the threshold explicitly and use the network for computing alignment score and not for the alignment itself.
For the hard alignment, we use $A_2^{0.001} \cap A_3^{1} \cap A_4^{1}$, which we found to work the best with this ensemble on the training data.

\paragraph{Additional Features.} Apart from $M_1$, $M_2^b$, $M_3^{aa}$, $M_3^{bb}$ and Attention with averaging aggregation (Individual), we also include the output of \fastalign{} as one of the features. Moreover, four other manually crafted features (Manual) are added. The motivation for the first two manual features is that the position and token length help in determining the alignment in some cases. The last two are specifically targeted at named entities, which have sparse occurrences in the data, and also at non-word tokens, such as full stops, delimiters and quotation marks. 
We list Pearson's correlation coefficient with true alignments on Czech$\leftrightarrow$English data (the two directions averaged).

\smallskip
\begin{itemize}
    \item Difference in sentence positions:\\
    $\rho = \text{-}\ 0.18$, \qquad $abs(\,i/|S|-j/|T|\,)$ 
    \item Difference in token lengths:\\
    $\rho = \text{-}\ 0.11$, \qquad $abs(\,|s_i|-|t_j|\,)$ 
    \item Difference in subword unit counts:\\
    $\rho = \text{-}\ 0.03$, \qquad $abs(\,|\text{subw}(s_i)|-|\text{subw}(t_j)|\,)$ 
    \item Normalized token case-insensitive Levenshtein distance:\\
    $\rho = \text{-}\ 0.30$, \qquad $lev(s_i, t_j)/max(|s_i|, |t_j|)$
    \item Number of subword units which are present in both tokens:\footnote{Normalized version of this feature had slightly lower correlation coefficient: $0.30$.}\\
    $\rho = \,\,\,0.32$, \qquad $|\,\text{subw}(s_i) \cap \text{subw}(t_j)\,|$
    \item Token string case-insensitive equality (equal to zero Levenshtein distance):\\
    $\rho = \,\,\,0.28$, \qquad $I_{s_i \simeq t_j}$
\end{itemize}

\paragraph{Architecture.} For every model, the epoch with the lowest AER on the validation dataset is used for the test dataset. This extractor was found to work best across all ensemble models. The training was done with cross-entropy loss. The model was composed of series of hidden linear layers, each with biases and Tanh as the activation function with dropouts around the innermost layer:
\begin{gather*}
    L_{|\text{Input}|}^\text{Tanh}\circ L_{32}^\text{Tanh} \circ D_{0.2} \circ L_{16}^\text{Tanh} \circ D_{0.2} \circ L_{16}^\text{Tanh} \circ L_{8}^\text{Tanh} \circ L_1^\text{Softmax}
\end{gather*}

\subsection{Data}

The Czech$\leftrightarrow$English dataset contains 1.5M source-target pairs, out of which $2.64\%$ is of a positive class (aligned). For German$\leftrightarrow$English Small these quantities are 22k and $5.61\%$ respectively. This could be an issue for a simple classifier network and would need e.g. oversampling of the positive or undersampling of the negative class.

For Czech$\leftrightarrow$English, we used $10\%$ and $10\%$ (250 sentences each) for validation and test data and the rest for training. Samples were split on sentence boundaries. The English$\rightarrow$German was used solely for testing, due to its small size.

\subsection{Evaluation}

The averaged results of each ensemble on Czech$\leftrightarrow$English are in \Cref{tab:ensemble_performance}. We also show the results of $M_1$, but without $A_2$. Due to the range of $M_1$'s values, it is difficult to establish a cut-off threshold. Attention uses $A_3^1$, since intersection with other extractors did not improve the performance, as described in \Cref{sec:evaluation}.
The results demonstrate that adding any feature improves the overall ensemble. All features combined together improve on the best individual model by $-0.11$\, AER.\footnote{Performed by Student's t-test on 10 runs with $p<0.001$.}

\newcommand{\staroff}{\hspace{0.02cm}$\star$\hspace*{-0.28cm}}
\begin{table*}[h!]
    \center
    \begin{tabular}{lccc}
        \toprule
        Model\,/\,Features & Precision & Recall & AER \\
        \midrule
        $M_1$ ($A_3^{1} \cap A_4^{1}$) & $0.75$ & $0.78$ & $0.25$ \staroff \\
        Attention (max, $A_3^1$) & $0.64$ & $0.81$ & $0.29$ \\
        \fastalign{} Small & $0.54$ & $0.66$ & $0.41$ \\
        \fastalign{} Big & $0.63$ & $0.64$ & $0.38$ \\
        \midrule
        Manual features & $0.55$ & $0.46$ & $0.50$ \\
        Individual ($M_1$, $M_2^b$, $M_3^{aa}$, $M_3^{bb}$, attention) &  $0.84$ & $0.73$ & $0.23$ \\
        Manual + Indiv. & $0.85$ & $0.79$ & $0.19$ \\
        Manual + Indiv. + \fastalign{} & $0.86$ & $0.79$ & $0.18$ \\
        Manual + Indiv. + \fastalign{} + Attention & $0.85$ & $0.84$ & $0.16$ \\
        Manual + Indiv. + \fastalign{} + Attention + M1 rev. \hspace*{-0.4cm} & $0.86$ & $0.86$ & $0.14$ \staroff \\
        \bottomrule
    \end{tabular}
    \caption{Average Precision, Recall and AER of $M_1$ (best individual) and different ensemble models (using $A_2^{0.001} \cap A_3^{1} \cap A_4^{1}$) on Czech$\leftrightarrow$English data (averaged) \label{tab:ensemble_performance}}
\end{table*}

\vspace{-0.2cm}
\paragraph{Transfer.} The best models on Czech$\leftrightarrow$English (one for each direction) were then used on the English$\rightarrow$German dataset, resulting on average in AER = $0.18$. This is higher than for Czech but still significantly lower (by a margin of $-0.08$)\footnote{Performed by Student's t-test with $p<0.001$.} than for the best individual model, Attention (max). This suggests that the features generalize well and models can be trained even on other language data. Furthermore, since the alignment datasets come from different origins, there may be systematic biases, which lower the performance of the transfer.
\section{Summary}
\vspace{-0.1cm}

This paper explored and compared different methods of inducing word alignment from trained NMT models.
Despite its simplicity, estimating scores with single word translations (combined with reverse translations) appears to be the fastest and the most robust solution, even compared to word alignment from attention heads.
Ensembling individual model scores with a simple feed-forward network improves the final performance to AER = $0.14$ on Czech$\leftrightarrow$English data.

\vspace{-0.3cm}
\paragraph{Future work.}

\Cref{subsec:baseline_models} presented but did not explore an idea of target dropout with multiple tokens in order to better model the fact that words rarely map 1:1.
We then used neural MT for providing alignment scores but then used a primitive extractor algorithm for obtaining hard alignment. More sophisticated approaches which consider the soft alignment origin (NMT), could vastly improve the performance.

Although it was possible to use any alignment extractor to get hard alignments out of soft ones, we found that the choice of the mechanism and also the parameters had a considerable influence on the performance. These alignment extractors are, however, not bound to alignment from NMT and their ability to be used with other soft alignment models and other symmetrization techniques should be examined further.

Finally, we did not explore the possible effects of fine-tuning the translation model on the available data or training it solely on this data. Similarity based on word embeddings could be used as yet another soft-alignment feature.

\section*{Acknowledgements}

We would like to thank Bernadeta Griciūtė and Nikola Kalábová for their reviews and insightful comments.

This article has used services provided by the LINDAT/CLARIAH-CZ Research Infrastructure (\href{https://lindat.cz}{lindat.cz}), supported by the Ministry of Education, Youth and Sports of the Czech Republic (Project No. LM2018101).
The work described also used trained models made available by the H2020-ICT-2018-2-825303 (Bergamot) grant.

\bibliography{mybib}

\begin{thebibliography}{32}
\providecommand{\natexlab}[1]{#1}
\providecommand{\url}[1]{\texttt{#1}}
\expandafter\ifx\csname urlstyle\endcsname\relax
  \providecommand{\doi}[1]{doi: #1}\else
  \providecommand{\doi}{doi: \begingroup \urlstyle{rm}\Url}\fi

\bibitem[Alkhouli et~al.(2016)Alkhouli, Bretschner, Peter, Hethnawi, Guta, and
  Ney]{alkhouli2016alignment}
Alkhouli, Tamer, Gabriel Bretschner, Jan-Thorsten Peter, Mohammed Hethnawi,
  Andreas Guta, and Hermann Ney.
\newblock Alignment-based neural machine translation.
\newblock In \emph{Proceedings of the First Conference on Machine Translation:
  Volume 1, Research Papers}, pages 54--65, 2016.

\bibitem[Bahdanau et~al.(2014)Bahdanau, Cho, and Bengio]{bahdanau2014neural}
Bahdanau, Dzmitry, Kyunghyun Cho, and Yoshua Bengio.
\newblock Neural machine translation by jointly learning to align and
  translate.
\newblock \emph{arXiv preprint arXiv:1409.0473}, 2014.

\bibitem[Bi\c{c}ici(2011)]{bicini_ende_algn_corpus}
Bi\c{c}ici, Ergun.
\newblock \emph{The Regression Model of Machine Translation}.
\newblock PhD thesis, Ko\c{c} University, 2011.
\newblock Supervisor: Deniz Yuret.

\bibitem[Bogoychev et~al.(2020)Bogoychev, Grundkiewicz, Aji, Behnke, Heafield,
  Kashyap, Farsarakis, and Chudyk]{model_deen}
Bogoychev, Nikolay, Roman Grundkiewicz, Alham~Fikri Aji, Maximiliana Behnke,
  Kenneth Heafield, Sidharth Kashyap, Emmanouil-Ioannis Farsarakis, and Mateusz
  Chudyk.
\newblock {E}dinburgh{'}s Submissions to the 2020 Machine Translation
  Efficiency Task.
\newblock In \emph{Proceedings of the Fourth Workshop on Neural Generation and
  Translation}, pages 218--224, Online, July 2020. Association for
  Computational Linguistics.
\newblock URL \url{https://www.aclweb.org/anthology/2020.ngt-1.26}.

\bibitem[Chen et~al.(2020{\natexlab{a}})Chen, Sun, and Liu]{chen2020mask}
Chen, Chi, Maosong Sun, and Yang Liu.
\newblock Mask-Align: Self-Supervised Neural Word Alignment.
\newblock \emph{arXiv preprint arXiv:2012.07162}, 2020{\natexlab{a}}.

\bibitem[Chen et~al.(2016)Chen, Matusov, Khadivi, and Peter]{chen2016guided}
Chen, Wenhu, Evgeny Matusov, Shahram Khadivi, and Jan-Thorsten Peter.
\newblock Guided alignment training for topic-aware neural machine translation.
\newblock \emph{arXiv preprint arXiv:1607.01628}, 2016.

\bibitem[Chen et~al.(2020{\natexlab{b}})Chen, Liu, Chen, Jiang, and
  Liu]{chen2020accurate}
Chen, Yun, Yang Liu, Guanhua Chen, Xin Jiang, and Qun Liu.
\newblock Accurate Word Alignment Induction from Neural Machine Translation.
\newblock \emph{arXiv preprint arXiv:2004.14837}, 2020{\natexlab{b}}.

\bibitem[Dyer et~al.(2013)Dyer, Chahuneau, and Smith]{dyer2013simple}
Dyer, Chris, Victor Chahuneau, and Noah~A Smith.
\newblock A simple, fast, and effective reparameterization of ibm model 2.
\newblock In \emph{Proceedings of the 2013 Conference of the North American
  Chapter of the Association for Computational Linguistics: Human Language
  Technologies}, pages 644--648, 2013.

\bibitem[Germann et~al.(2020)Germann, Grundkiewicz, Popel, Dobreva, Bogoychev,
  and Heafield]{model_csen}
Germann, Ulrich, Roman Grundkiewicz, Martin Popel, Radina Dobreva, Nikolay
  Bogoychev, and Kenneth Heafield.
\newblock Speed-optimized, Compact Student Models that Distill Knowledge from a
  Larger Teacher Model: the UEDIN-CUNI Submission to the WMT 2020 News
  Translation Task.
\newblock In \emph{Proceedings of the Fifth Conference on Machine Translation},
  pages 190--195, Online, Nov. 2020. Association for Computational Linguistics.
\newblock URL \url{https://www.aclweb.org/anthology/2020.wmt-1.17}.

\bibitem[Junczys-Dowmunt and Sza{\l}(2011)]{junczys2011symgiza++}
Junczys-Dowmunt, Marcin and Arkadiusz Sza{\l}.
\newblock Symgiza++: symmetrized word alignment models for statistical machine
  translation.
\newblock In \emph{International Joint Conferences on Security and Intelligent
  Information Systems}, pages 379--390. Springer, 2011.

\bibitem[Junczys-Dowmunt et~al.(2018{\natexlab{a}})Junczys-Dowmunt,
  Grundkiewicz, Dwojak, Hoang, Heafield, Neckermann, Seide, Germann, Aji,
  Bogoychev, et~al.]{junczys2018marian}
Junczys-Dowmunt, Marcin, Roman Grundkiewicz, Tomasz Dwojak, Hieu Hoang, Kenneth
  Heafield, Tom Neckermann, Frank Seide, Ulrich Germann, Alham~Fikri Aji,
  Nikolay Bogoychev, et~al.
\newblock Marian: Fast neural machine translation in {C}++.
\newblock \emph{arXiv preprint arXiv:1804.00344}, 2018{\natexlab{a}}.

\bibitem[Junczys-Dowmunt et~al.(2018{\natexlab{b}})Junczys-Dowmunt, Heafield,
  Hoang, Grundkiewicz, and Aue]{junczys2018marian2}
Junczys-Dowmunt, Marcin, Kenneth Heafield, Hieu Hoang, Roman Grundkiewicz, and
  Anthony Aue.
\newblock Marian: Cost-effective high-quality neural machine translation in
  C++.
\newblock \emph{arXiv preprint arXiv:1805.12096}, 2018{\natexlab{b}}.

\bibitem[Kocmi et~al.(2020)Kocmi, Popel, and Bojar]{czeng2}
Kocmi, Tom, Martin Popel, and Ondrej Bojar.
\newblock Announcing czeng 2.0 parallel corpus with over 2 gigawords.
\newblock \emph{arXiv preprint arXiv:2007.03006}, 2020.

\bibitem[Koehn(2009)]{koehn2009statistical}
Koehn, Philipp.
\newblock \emph{Statistical machine translation}.
\newblock Cambridge University Press, 2009.

\bibitem[Kudo and Richardson(2018)]{kudo2018sentencepiece}
Kudo, Taku and John Richardson.
\newblock Sentencepiece: A simple and language independent subword tokenizer
  and detokenizer for neural text processing.
\newblock \emph{arXiv preprint arXiv:1808.06226}, 2018.

\bibitem[Li et~al.(2019)Li, Li, Liu, Meng, and Shi]{li2019word}
Li, Xintong, Guanlin Li, Lemao Liu, Max Meng, and Shuming Shi.
\newblock On the word alignment from neural machine translation.
\newblock In \emph{Proceedings of the 57th Annual Meeting of the Association
  for Computational Linguistics}, pages 1293--1303, 2019.

\bibitem[Liang et~al.(2006)Liang, Taskar, and Klein]{liang2006alignment}
Liang, Percy, Ben Taskar, and Dan Klein.
\newblock Alignment by agreement.
\newblock In \emph{Proceedings of the Human Language Technology Conference of
  the NAACL, Main Conference}, pages 104--111, 2006.

\bibitem[Mare{\v c}ek(2016)]{marecek_csen_algn_corpus}
Mare{\v c}ek, David.
\newblock Czech-English Manual Word Alignment, 2016.
\newblock URL \url{http://hdl.handle.net/11234/1-1804}.
\newblock {LINDAT}/{CLARIAH}-{CZ} digital library at the Institute of Formal
  and Applied Linguistics ({{\'U}FAL}), Faculty of Mathematics and Physics,
  Charles University.

\bibitem[Matusov et~al.(2004)Matusov, Zens, and Ney]{symmetric}
Matusov, Evgeny, Richard Zens, and Hermann Ney.
\newblock Symmetric word alignments for statistical machine translation.
\newblock 01 2004.
\newblock \doi{10.3115/1220355.1220387}.

\bibitem[Mihalcea and Pedersen(2003)]{mihalcea2003evaluation}
Mihalcea, Rada and Ted Pedersen.
\newblock An evaluation exercise for word alignment.
\newblock In \emph{Proceedings of the HLT-NAACL 2003 Workshop on Building and
  using parallel texts: data driven machine translation and beyond}, pages
  1--10, 2003.

\bibitem[Och and Ney(2000)]{och2000improved}
Och, Franz~Josef and Hermann Ney.
\newblock Improved statistical alignment models.
\newblock In \emph{Proceedings of the 38th annual meeting of the association
  for computational linguistics}, pages 440--447, 2000.

\bibitem[Och and Ney(2003)]{och2003systematic}
Och, Franz~Josef and Hermann Ney.
\newblock A systematic comparison of various statistical alignment models.
\newblock \emph{Computational linguistics}, 29\penalty0 (1):\penalty0 19--51,
  2003.

\bibitem[Post(2018)]{sacrebleu}
Post, Matt.
\newblock A Call for Clarity in Reporting {BLEU} Scores.
\newblock In \emph{Proceedings of the Third Conference on Machine Translation:
  Research Papers}, pages 186--191, Belgium, Brussels, Oct. 2018. Association
  for Computational Linguistics.
\newblock URL \url{https://www.aclweb.org/anthology/W18-6319}.

\bibitem[Rozis and Skadi{\c{n}}{\v{s}}(2017)]{rozis_tilde}
Rozis, Roberts and Raivis Skadi{\c{n}}{\v{s}}.
\newblock Tilde MODEL-multilingual open data for EU languages.
\newblock In \emph{Proceedings of the 21st Nordic Conference on Computational
  Linguistics}, pages 263--265, 2017.

\bibitem[Sabet et~al.(2020)Sabet, Dufter, and Sch{\"u}tze]{sabet2020simalign}
Sabet, Masoud~Jalili, Philipp Dufter, and Hinrich Sch{\"u}tze.
\newblock Simalign: High quality word alignments without parallel training data
  using static and contextualized embeddings.
\newblock \emph{arXiv preprint arXiv:2004.08728}, 2020.

\bibitem[Schrader(2006)]{schrader2006does}
Schrader, Bettina.
\newblock How does morphological complexity translate? A cross-linguistic case
  study for word alignment.
\newblock In \emph{Proceedings of Linguistic Evidence Conference}, pages
  189--191, 2006.

\bibitem[Specia et~al.(2013)Specia, Shah, De~Souza, and Cohn]{specia2013quest}
Specia, Lucia, Kashif Shah, Jos{\'e}~GC De~Souza, and Trevor Cohn.
\newblock QuEst-A translation quality estimation framework.
\newblock In \emph{Proceedings of the 51st Annual Meeting of the Association
  for Computational Linguistics: System Demonstrations}, pages 79--84, 2013.

\bibitem[Vaswani et~al.(2017)Vaswani, Shazeer, Parmar, Uszkoreit, Jones, Gomez,
  Kaiser, and Polosukhin]{vaswani2017transformer}
Vaswani, Ashish, Noam Shazeer, Niki Parmar, Jakob Uszkoreit, Llion Jones,
  Aidan~N. Gomez, Lukasz Kaiser, and Illia Polosukhin.
\newblock Attention Is All You Need, 2017.

\bibitem[Wu et~al.(2021)Wu, Ding, Yang, and Tao]{wu2021slua}
Wu, Di, Liang Ding, Shuo Yang, and Dacheng Tao.
\newblock SLUA: A Super Lightweight Unsupervised Word Alignment Model via
  Cross-Lingual Contrastive Learning.
\newblock \emph{arXiv preprint arXiv:2102.04009}, 2021.

\bibitem[Zenkel et~al.(2019)Zenkel, Wuebker, and DeNero]{zenkel2019adding}
Zenkel, Thomas, Joern Wuebker, and John DeNero.
\newblock Adding interpretable attention to neural translation models improves
  word alignment.
\newblock \emph{arXiv preprint arXiv:1901.11359}, 2019.

\bibitem[Zintgraf et~al.(2017)Zintgraf, Cohen, Adel, and
  Welling]{zintgraf2017visualizing}
Zintgraf, Luisa~M, Taco~S Cohen, Tameem Adel, and Max Welling.
\newblock Visualizing deep neural network decisions: Prediction difference
  analysis.
\newblock \emph{arXiv preprint arXiv:1702.04595}, 2017.

\bibitem[Zouhar and Nov{\'a}k(2020)]{zouhar2020extending}
Zouhar, Vil{\'e}m and Michal Nov{\'a}k.
\newblock Extending {P}takop{\v{e}}t for Machine Translation User Interaction
  Experiments.
\newblock \emph{The Prague Bulletin of Mathematical Linguistics}, \penalty0
  (115):\penalty0 129--142, 2020.

\end{thebibliography}


\correspondingaddress
\end{document}